\documentclass{article}

\usepackage[utf8]{inputenc}
\usepackage[T1]{fontenc}

\usepackage[preprint]{icml2026}
\usepackage{hyperref}

\usepackage{booktabs}
\usepackage{amsfonts}
\usepackage{amsmath}
\usepackage{amssymb}
\usepackage{nicefrac}
\usepackage{microtype}
\usepackage{tikz}
\usepackage{enumitem}
\usetikzlibrary{positioning,arrows.meta,fit,calc,backgrounds}

\icmltitlerunning{SmartSearch: How Ranking Beats Structure for Conversational Memory Retrieval}

\begin{document}

\twocolumn[
\icmltitle{SmartSearch: How Ranking Beats Structure for\\
Conversational Memory Retrieval}

\icmlsetsymbol{equal}{*}

\begin{icmlauthorlist}
\icmlauthor{Jesper Derehag}{midbrain}
\icmlauthor{Carlos Calva}{midbrain}
\icmlauthor{Timmy Ghiurau}{midbrain}
\end{icmlauthorlist}

\icmlaffiliation{midbrain}{Midbrain}

\icmlcorrespondingauthor{Jesper Derehag}{jesper@midbrain.ai}

\printAffiliationsAndNotice{}
]

{\let\thefootnote\relax\footnotetext{\hspace*{-\footnotesep}%
\textsuperscript{1}\,Midbrain. Correspondence to: Jesper Derehag
\textless{}jesper@midbrain.ai\textgreater{}.}}

\begin{abstract}
Recent conversational memory systems invest heavily in LLM-based structuring at ingestion time and learned retrieval policies at query time. We show that neither is necessary. SmartSearch retrieves from raw, unstructured conversation history using a fully deterministic pipeline: NER-weighted substring matching for recall, rule-based entity discovery for multi-hop expansion, and a CrossEncoder+ColBERT rank fusion stage---the only learned component---running on CPU in ${\sim}$650\,ms. Oracle analysis on two benchmarks identifies a \emph{compilation bottleneck}: retrieval recall reaches 98.6\%, but without intelligent ranking only 22.5\% of gold evidence survives truncation to the token budget. With score-adaptive truncation and no per-dataset tuning, SmartSearch achieves 93.5\% on LoCoMo and 88.4\% on LongMemEval-S, exceeding all known memory systems under the same evaluation protocol on both benchmarks while using 8.5$\times$ fewer tokens than full-context baselines.
\end{abstract}

\section{Introduction}

Large language model agents that maintain conversations over days or weeks face a fundamental challenge: context windows, even at 100K{+} tokens~\citep{anthropic2024,reid2024}, cannot encompass the full history of extended interactions. Recent memory systems address this through increasingly sophisticated architectures that deploy LLMs at two stages:

\begin{enumerate}
\item \textbf{Memory structuring (ingestion time):} Before any query arrives, systems use LLMs or learned models to reorganize raw conversation into structured representations---consolidating episodes into abstract MemCells~\citep{evermemos2026}, clustering traces into narrative themes~\citep{shu2026}, maintaining dual-layer abstractions~\citep{xia2026}, or learning skill-based memory operations~\citep{zhang2026memskill}. This preprocessing is computationally expensive, but its cost is amortized and rarely reported alongside per-query token counts, making systems appear cheaper at query time than they are in total.
\item \textbf{Query-time retrieval:} At query time, many systems use LLMs to generate search queries, route across memory tiers, or iteratively refine results~\citep{lewis2020,gao2024}. Each LLM call adds substantial latency and cost. Learned routing policies~\citep{zhang2026budgetmem} require training data and GPU inference during search.
\end{enumerate}

SmartSearch does neither. It operates on raw, unstructured conversation history---no consolidation, no clustering, no abstraction layers---and performs search with no LLM in the loop. The motivation is a simple observation: when LLM agents are given shell access to search a filesystem (the setup that produces the strongest memory retrieval results), their search behavior is overwhelmingly deterministic---extract entities from the query, search for substrings, read the results. If the search strategy is deterministic, why pay for an LLM to execute it?

We test this hypothesis directly. SmartSearch's entire retrieval pipeline is deterministic at search time:

\begin{enumerate}
\item \textbf{Query understanding without LLMs:} SpaCy NER/POS tagging extracts and weights search terms directly from the query. No query generation, no learned embeddings, no API calls.
\item \textbf{Multi-hop expansion without learned policies:} When the first retrieval hop returns results, rule-based NER discovers new entities (e.g., a person's name mentioned in a retrieved passage), which seed subsequent hops. No reinforcement learning, no trained router---just deterministic entity extraction on retrieved text.
\item \textbf{Retrieval without GPU:} NER-weighted substring matching handles 98.9\% of oracle traces, running on CPU in milliseconds.
\end{enumerate}

The only ML component is a CrossEncoder reranker (\texttt{mxbai-rerank-large-v1}, 435M parameters, DeBERTaV3) fused with ColBERT~\citep{khattab2020} via Reciprocal Rank Fusion (RRF), applied \emph{after} retrieval is complete to prioritize passages before token-budget truncation. Both models are small enough to run on CPU; since they score independently, they execute in parallel with wall-clock latency ${\sim}$650\,ms per query---no GPU required.

On the LoCoMo benchmark, this architecture achieves 91.9\% accuracy with 3,141 average tokens---8.5$\times$ fewer than full-context baselines while scoring 14.8\,pp higher. Oracle analysis reveals why: raw retrieval recall is already 98.6\%, but without intelligent ranking only 22.5\% of gold evidence survives truncation to the token budget. The reranker bridges this \emph{compilation bottleneck}.

We make three contributions:

\begin{enumerate}
\item \textbf{An LLM-free multi-hop retrieval architecture} that uses NER-weighted substring matching for recall, rule-based entity discovery for hop expansion, and a lightweight CrossEncoder+ColBERT rank fusion stage as the only learned component. Both rerankers are small enough to run on CPU and score independently, enabling parallel execution (${\sim}$650\,ms wall-clock, no GPU). Linguistic term weighting (POS/NER) replaces corpus-statistical weighting (BM25/IDF), and all evidence is collected in a single pass before any model inference. We further show that the oracle-derived insight---that substring matching resolves 98.9\% of queries and learned tool routing is degenerate---enables a fully \emph{index-free} variant that drops all precomputed indices (Section~\ref{sec:expansion}), operating on raw text files with \texttt{grep} as the sole retrieval primitive.
\item \textbf{Empirical validation on two benchmarks} showing 91.9\% accuracy on LoCoMo~\citep{maharana2024} (${\sim}$9K-token conversations) with 8.5$\times$ token efficiency versus full-context baselines, and competitive results on LongMemEval-S~\citep{wu2024longmemeval} (${\sim}$115K-token conversations), where the index-free variant with query expansion exceeds the indexed baseline. Systematic ablation across 27 configurations on LoCoMo traces the 7.2\,pp improvement from baseline to final system.
\item \textbf{A compilation bottleneck analysis} demonstrating that the primary accuracy limiter is not retrieval quality but post-retrieval ranking---an insight confirmed on both short and long conversations, explaining why sophisticated search policies yield diminishing returns while query expansion and reranker quality yield large gains.
\end{enumerate}

\section{Related Work}

\subsection{Long-Context Memory Systems}

A defining feature of recent memory systems is that they invest significant computation \emph{before} any query arrives, using LLMs or learned models to restructure raw conversation into representations optimized for retrieval. This preprocessing cost is amortized across queries and typically unreported, so systems that appear to use few tokens at query time may rely on expensive ingestion pipelines.

\textbf{LLM-structured memory:} EverMemOS~\citep{evermemos2026} uses an LLM agent to consolidate episodic memory into MemCells (atomic facts) and MemScenes (episode summaries), then performs agentic multi-round retrieval (BM25 + vector + exact match) at query time---achieving 92.3\% on LoCoMo. Memora~\citep{xia2026} maintains a dual-layer abstraction: LLM-generated abstract indices for routing and concrete memory values for detail. M2A~\citep{feng2026} combines raw message logs with LLM-generated semantic memory. In each case, the LLM does the heavy lifting of deciding \emph{what is worth remembering and how to organize it}---a form of pre-query intelligence that is invisible in per-query token counts.

\textbf{ML-structured memory:} TraceMem~\citep{shu2026} applies hierarchical clustering to organize conversation traces into topological themes. MemSkill~\citep{zhang2026memskill} learns to select and evolve memory skills (extraction, consolidation, pruning). BudgetMem~\citep{zhang2026budgetmem} learns routing policies across tiered memory modules with reinforcement learning. Locas~\citep{lu2026} attaches parameter-efficient memory to FFN blocks. These approaches use smaller ML models rather than full LLMs for structuring, but still require task-specific training data and GPU inference during ingestion or search.

\textbf{Parametric memory:} LongMem~\citep{wang2023} augments LLMs with learnable memory modules that absorb conversation into model parameters. These approaches require careful regularization to prevent catastrophic forgetting~\citep{kirkpatrick2017}.

\textbf{SmartSearch's position:} SmartSearch operates on raw, unstructured conversation with no ingestion-time processing. The entire retrieval loop is deterministic; the only learned component is a CrossEncoder+ColBERT reranker applied \emph{after} retrieval to prioritize passages before truncation. The question is whether careful search and ranking over raw text can match systems that invest in structuring the memory itself.

\subsection{Retrieval Methods}

\textbf{Keyword search:} Substring-based exact matching is fast but brittle. BM25~\citep{robertson2009} provides probabilistic ranking and is the standard for keyword retrieval.

\textbf{Dense retrieval:} Embedding-based methods such as DPR~\citep{karpukhin2020} and ColBERT~\citep{khattab2020} bridge vocabulary gaps but are slower and require pre-computed indices. Recent work~\citep{li2026} shows that listwise reranking with lightweight models outperforms pointwise scoring.

\textbf{Hybrid approaches:} Recent systems combine keyword and dense retrieval~\citep{bruch2023,luan2021}. SmartSearch follows this trend with substring matching for recall and ColBERT for the vocabulary-gap tail, but replaces BM25's corpus-statistical term weighting with linguistic weighting via NER/POS.

\subsection{Multi-Hop Retrieval and Question Answering}

Multi-hop reasoning has been studied extensively in open-domain QA. HotpotQA~\citep{yang2018} established the benchmark for multi-hop questions requiring reasoning over multiple documents. Subsequent work developed iterative retrieval strategies: IRCoT~\citep{trivedi2023} interleaves chain-of-thought reasoning with retrieval, using the LLM to generate intermediate queries. ITER-RETGEN~\citep{shao2023} iteratively retrieves and generates to refine answers. Self-Ask~\citep{press2023} decomposes complex questions into sub-questions answered sequentially. ReAct~\citep{yao2023} combines reasoning traces with action steps for multi-hop retrieval.

These approaches share a common pattern: they use the LLM as a query generator within the retrieval loop, effectively giving the LLM shell-like access to a search backend. Yet examining what the LLM actually \emph{does} in this loop---extract entities from the query, search for them, read results, extract new entities---reveals that the behavior is largely deterministic. SmartSearch replaces the LLM with rule-based NER over retrieved text, achieving multi-hop expansion at zero inference cost. Oracle analysis (Section~\ref{sec:oracle}) confirms this suffices: 97\% of LoCoMo queries resolve in a single hop.

\subsection{Retrieval-Augmented Generation}

RAG~\citep{lewis2020} established the paradigm of augmenting LLMs with retrieved evidence. DPR~\citep{karpukhin2020} introduced dense passage retrieval using dual encoders. Subsequent work has scaled RAG to long-context settings~\citep{gao2024}, introduced more sophisticated retrieval strategies~\citep{asai2024}, and explored active retrieval where the model decides when to retrieve~\citep{jiang2023}. SmartSearch operates within the RAG paradigm but replaces the neural retriever with deterministic tools.

\subsection{Reranking}

\textbf{Cross-encoder reranking:} Models like ms-marco-MiniLM-L-12-v2~\citep{reimers2019} score query-document pairs directly, achieving state-of-the-art ranking quality. These models are fine-tuned on MS~MARCO~\citep{nguyen2016}, a large-scale passage ranking dataset. \citet{li2026} show that ranking plays a critical role in long-context retrieval, validating SmartSearch's use of reranking as a core component.

\textbf{Listwise ranking:} Recent work~\citep{li2026} shows that ranking across entire candidate sets (listwise) outperforms pointwise ranking for long-context dialogue.

\section{Method}

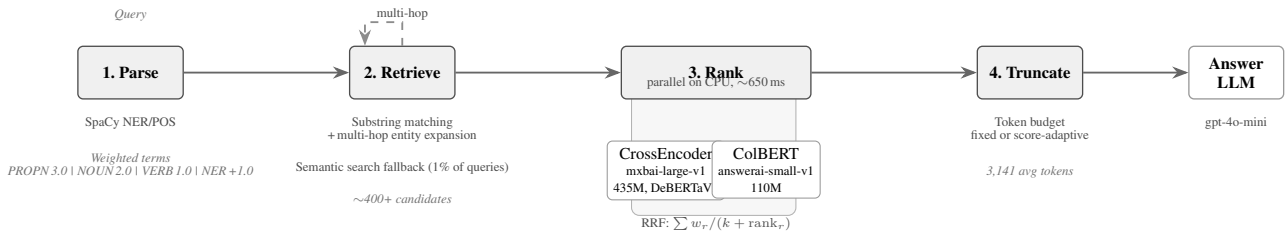
\begin{figure*}[t]
\centering
\begin{tikzpicture}[
    >=Stealth,
    node distance=0.6cm and 0.7cm,
    stage/.style={
        rectangle, draw=black!70, fill=black!6,
        minimum height=1.0cm, minimum width=2.0cm,
        align=center, font=\small\bfseries,
        rounded corners=2pt, line width=0.5pt
    },
    substage/.style={
        rectangle, draw=black!40, fill=white,
        minimum height=0.65cm, minimum width=1.8cm,
        align=center, font=\footnotesize,
        rounded corners=2pt, line width=0.4pt
    },
    annot/.style={font=\scriptsize, text=black!70, align=center},
    dataarrow/.style={->, thick, black!60, line width=0.7pt},
    io/.style={font=\scriptsize\itshape, text=black!55, align=center},
    scale=0.70, every node/.style={scale=0.70}
]

\node[stage] (parse) {1.~Parse};
\node[annot, below=0.15cm of parse] (parse-detail) {SpaCy NER/POS};
\node[io, above=0.25cm of parse] (parse-in) {Query};
\node[io, below=0.15cm of parse-detail] (parse-out) {Weighted terms\\[-1pt]{\scriptsize PROPN\,3.0 \textbar{} NOUN\,2.0 \textbar{} VERB\,1.0 \textbar{} NER\,+1.0}};

\node[stage, right=2.2cm of parse] (retrieve) {2.~Retrieve};
\node[annot, below=0.15cm of retrieve] (ret-detail) {Substring matching\\[-1pt]{\scriptsize +\,multi-hop entity expansion}};
\node[annot, below=0.1cm of ret-detail] (ret-fallback) {\scriptsize Semantic search fallback (1\% of queries)};
\node[io, below=0.1cm of ret-fallback] (ret-out) {${\sim}$400+ candidates};

\node[stage, right=2.2cm of retrieve, minimum width=3.6cm] (rank) {3.~Rank};

\node[substage, below=0.55cm of rank, xshift=-0.95cm] (ce) {CrossEncoder\\[-1pt]{\scriptsize mxbai-large-v1}\\[-1pt]{\scriptsize 435M, DeBERTaV3}};
\node[substage, below=0.55cm of rank, xshift=0.95cm] (cb) {ColBERT\\[-1pt]{\scriptsize answerai-small-v1}\\[-1pt]{\scriptsize 110M}};

\node[annot, above=0.06cm of rank.south] (parallel-label) {\scriptsize parallel on CPU, ${\sim}$650\,ms};

\node[annot, below=0.15cm of $(ce.south)!0.5!(cb.south)$] (rrf) {RRF: $\sum w_r / (k + \mathrm{rank}_r)$};

\begin{scope}[on background layer]
\node[draw=black!30, fill=black!3, rounded corners=3pt, line width=0.4pt,
      fit=(rank)(ce)(cb)(rrf)(parallel-label),
      inner sep=4pt] (rankbox) {};
\end{scope}

\node[stage, right=2.2cm of rank] (trunc) {4.~Truncate};
\node[annot, below=0.15cm of trunc] (trunc-detail) {Token budget\\[-1pt]{\scriptsize fixed or score-adaptive}};
\node[io, below=0.15cm of trunc-detail] (trunc-out) {3{,}141 avg tokens};

\node[stage, right=1.4cm of trunc, draw=black!40, fill=white,
      minimum width=1.8cm] (llm) {Answer\\LLM};
\node[annot, below=0.15cm of llm] (llm-detail) {\scriptsize gpt-4o-mini};

\draw[dataarrow] (parse) -- (retrieve);
\draw[dataarrow] (retrieve) -- (rank);
\draw[dataarrow] (rank) -- (trunc);
\draw[dataarrow] (trunc) -- (llm);

\draw[dataarrow, dashed, black!45] (retrieve.north) -- ++(0,0.45)
    node[annot, above, yshift=-0.08cm] {\scriptsize multi-hop}
    -| ([xshift=0.3cm]retrieve.north west);

\end{tikzpicture}
\caption{SmartSearch pipeline. Stages 1--2 are fully deterministic; Stage~3 (Rank) is the only learned component, running two models in parallel on CPU. Dashed arrow: optional multi-hop entity expansion (3\% of queries).}
\label{fig:pipeline}
\end{figure*}

SmartSearch is a retrieve-then-rank pipeline that operates on raw conversation text with no LLM in the retrieval loop (Figure~\ref{fig:pipeline}). The design has four stages: (1)~parse the query into NER/POS-weighted terms, (2)~retrieve candidates via substring matching with optional multi-hop expansion, (3)~rank via CrossEncoder+ColBERT fusion, and (4)~truncate to token budget. Stages 1--2 are fully deterministic; Stage~3 is the only learned component.

\subsection{Evaluation}
\label{sec:eval-protocol}

We evaluate on two benchmarks: \textbf{LoCoMo-10}~\citep{maharana2024} (a 10-conversation, 1,540-question subset of the full 50-conversation LoCoMo dataset, ${\sim}$9K tokens per conversation, 4 non-adversarial question categories) and \textbf{LongMemEval-S}~\citep{wu2024longmemeval} (500 questions across ${\sim}$115K-token conversation histories, 6 of 7 question types excluding abstention). LoCoMo provides passage-level gold evidence; on LongMemEval-S, per-passage gold labels were derived from per-turn answer annotations (479 questions with gold, median 2 gold passages). Unless noted, LoCoMo uses gpt-4o-mini as both answer LLM and judge (binary J-score); LongMemEval-S ablation studies use Claude Sonnet 4.6 as answer LLM and gpt-4o-mini as judge. Main results (Table~\ref{tab:lme-results}) use gpt-4.1-mini as answer LLM and gpt-4o-mini as judge, matching the protocol of \citet{evermemos2026} and \citet{xia2026}.

\textbf{Cross-framework alignment.} Published LoCoMo numbers are not directly comparable because evaluation frameworks differ along multiple axes~\citep{zheng2023,wang2024pandalm}: (i)~answer LLM (gpt-4o-mini vs.\ gpt-4.1-mini), (ii)~judge model and prompt (binary J-score vs.\ 7-step chain-of-thought rubric), (iii)~answer prompt (direct vs.\ CoT), and (iv)~dataset split (LoCoMo-10 vs.\ LoCoMo-5). The full-context baseline alone shifts from 77.1\% to 91.2\% across frameworks---a 14\,pp swing with no retrieval change. Given these discrepancies, we report LoCoMo results under both our protocol and EverMemOS's~\citep{evermemos2026} protocol separately and never compare numbers across them.

\subsection{Oracle Trace Derivation}
\label{sec:oracle}

To determine the minimum-cost retrieval strategy for each question, we model the retrieval space as a directed graph where nodes are search states and edges are (tool, term-subset) actions with unit cost. Dijkstra's algorithm yields the shortest path covering the gold evidence passages $G_q$---the optimal trace. The oracle never sees gold \emph{answers}, only gold \emph{evidence passages}, preventing answer leakage. We run oracle derivation on LoCoMo~\citep{maharana2024}, where passage-level gold annotations are available for 1,536 questions (1,317 successful traces, 85.7\%).

Table~\ref{tab:oracle} summarizes the properties that motivate the pipeline design: retrieval is single-hop and grep-dominated, and the accuracy limiter is ranking, not search. The ranking effect generalizes across both benchmarks.

\begin{table}[t]
\centering
\caption{Oracle trace analysis (LoCoMo, 1,317 traces) and ranking effect on both benchmarks. Mean gold rank: average position of the first gold passage among grep candidates.}
\label{tab:oracle}
\vskip 0.1in
\small
\begin{tabular}{@{}lcc@{}}
\toprule
\textbf{Property} & \textbf{LoCoMo} & \textbf{LME-S} \\
\midrule
\multicolumn{3}{@{}l}{\textit{Hop distribution (oracle, LoCoMo only)}} \\
~~Single-hop & 97.0\% & --- \\
~~Two-hop & 2.8\% & --- \\
~~Three-hop & 0.2\% & --- \\
\midrule
\multicolumn{3}{@{}l}{\textit{Tool selection (oracle, LoCoMo only)}} \\
~~Resolved by grep & 98.9\% & --- \\
~~Resolved by semantic search & 1.0\% & --- \\
~~Resolved by grep (AND mode) & 0.1\% & --- \\
\midrule
\multicolumn{3}{@{}l}{\textit{Ranking effect}} \\
~~Total passages & 601 & 494 \\
~~Grep candidates & 431 & 120 \\
~~Mean gold rank (no reranker) & 195 & 47 \\
~~Mean gold rank (with CE) & 8 & 2 \\
\bottomrule
\end{tabular}
\end{table}

\subsection{Parse and Retrieve}

\textbf{Parse.} SpaCy \texttt{en\_core\_web\_sm}~\citep{honnibal2020} extracts terms via POS tagging and NER. Terms are weighted by linguistic specificity: proper nouns (3.0) $>$ nouns (2.0) $>$ verbs (1.0), with a +1.0 bonus for named entities. This replaces BM25's corpus-statistical weighting (IDF) with a linguistic signal: a speaker name appearing frequently in the conversation (low IDF) receives the highest weight because it is a proper noun and named entity.

\textbf{Retrieve.}
The recall stage uses exact substring matching (\texttt{grep}) over the NER-weighted terms. Substring matching has perfect recall for any passage containing the query term; precision is deferred entirely to ranking. For the 3\% of queries requiring multiple hops, entity expansion runs SpaCy NER on retrieved passages to discover entities absent from the original query, then searches again with the new terms. A dense retrieval model serves as a fallback for the ${\sim}$1\% of queries where substring matching fails due to vocabulary gaps (we use ColBERT~\citep{khattab2020}, but any embedding model filling this slot would suffice).

\subsection{Query Expansion}
\label{sec:expansion}

Grep's limitation is exact-match: when the query uses ``\emph{run}'' and the corpus says ``\emph{ran},'' grep misses the match. Rather than morphological expansion (tested and rejected---noisy candidates cascade through multi-hop), SmartSearch uses two expansion mechanisms:

\begin{itemize}
\item \textbf{Entity discovery:} After each hop, SpaCy NER extracts person, organization, location, and event names from retrieved passages. New entities not in the original query become search terms (weight 2.5) for subsequent hops.
\item \textbf{Pseudo-relevance feedback (PRF):} After main search hops, a PRF hop extracts frequent content words (nouns and proper nouns in $\geq$2 of the top-10 passages) and runs one additional \texttt{grep} pass with low match weight (0.5).
\end{itemize}

These mechanisms produce an \emph{index-free} variant that drops semantic retrieval entirely---no embedding index, no inverted index---leaving \texttt{grep} as the sole retrieval primitive and a single CrossEncoder as the sole ranker.

\subsection{Rank}

Without reranking, the first gold passage sits at mean rank 195 (LoCoMo) or 47 (LME-S) among grep candidates---far beyond any reasonable token budget. CE reranking pushes gold to rank 8 and 2 respectively (Table~\ref{tab:oracle}). This \emph{compilation bottleneck} motivates ranking as the core learned component.

The candidate set is scored by two complementary models running in parallel on CPU:

\begin{itemize}[nosep]
\item \textbf{CrossEncoder:} \texttt{mxbai-rerank-large-v1} (435M, DeBERTaV3~\citep{he2021}), fine-tuned on MS~MARCO~\citep{nguyen2016}. Pointwise scoring via Sentence-Transformers~\citep{reimers2019}.
\item \textbf{ColBERT:} Late-interaction scoring~\citep{khattab2020}. Spearman $\rho$ of 0.19--0.62 with the CrossEncoder, indicating complementary failure modes.
\end{itemize}

\noindent Rankings are fused via Reciprocal Rank Fusion: $\text{RRF}(d) = \sum_{r} w_r / (k + \text{rank}_r(d))$ with $k{=}60$, $w_{\text{CE}}{=}0.7$, $w_{\text{CB}}{=}0.3$. Wall-clock time is $\max(\text{CE}, \text{ColBERT}) \approx 650$\,ms on CPU; no GPU required.

\subsection{Truncate}

The fused ranking must be truncated before the answer LLM sees it. The simplest strategy is a \emph{fixed word budget}: include passages in rank order until a word limit is exhausted (default 2,000 words, yielding 3,141 avg tokens on LoCoMo). This works well on short conversations where the budget covers most of the ranked list, but becomes a binding constraint on longer corpora: on LongMemEval-S (${\sim}$115K tokens), the budget fills entirely and session diversity---critical for temporal and multi-session questions---is limited by the fixed cutoff.

An alternative is \emph{score-adaptive truncation}. Given the CrossEncoder scores from the ranking stage, we define a fractional threshold $\tau = \alpha \cdot \max_d(\text{CE}(q,d))$, where $\alpha \in (0,1)$ is a hyperparameter. Passages scoring below $\tau$ are pruned before the budget is applied. The intuition is query-adaptive allocation: queries with a strong top match and steep score drop-off (easy queries) receive compact context, while queries with flat score curves (hard queries) retain more passages. To avoid over-pruning from long-tailed candidate sets, the threshold is applied after a top-$K$ pre-selection step (e.g., $K{=}60$ by RRF score), restricting pruning to the high-scoring head.

The choice between fixed-budget and score-adaptive truncation is a precision--coverage tradeoff that interacts with corpus scale. On short conversations, the fixed budget is near-optimal because few low-relevance passages enter the context. On longer conversations, score-adaptive truncation with a raised budget ceiling (e.g., 3,500 words) can improve session diversity without proportionally increasing noise. We evaluate both strategies in Section~\ref{sec:truncation-results}.

\subsection{Experimental Setup}

We evaluate SmartSearch predominantly through ablation: starting from the full pipeline, we isolate each stage's contribution by removing or substituting components. Table~\ref{tab:pipeline-families} summarizes the configurations compared in Section~\ref{sec:experiments}. All share the same Parse stage (SpaCy NER/POS weighting). The indexed pipeline uses ColBERT for both retrieval and ranking; the index-free variants drop ColBERT entirely, compensating with query expansion to broaden grep recall.

\begin{table}[t]
\centering
\caption{Ablation structure and naming scheme. Configurations are composites of retrieval, rank, and budget choices (e.g., \texttt{I0-R5-B0}). Section~\ref{sec:experiments} tables vary one dimension and hold the others constant.}
\label{tab:pipeline-families}
\vskip 0.1in
\small
\begin{tabular}{@{}lcc@{}}
\toprule
\textbf{Stage} & \textbf{Indexed} & \textbf{Index-free} \\
\midrule
Retrieve & I\textit{n} & IF\textit{n} \\
\midrule
\multicolumn{3}{@{}l}{\textbf{Rank}} \\
~~No reranker & R0 & \\
~~\dots & & \\
~~CrossEncoder + RRF & R\textit{n} & \\
\midrule
\multicolumn{3}{@{}l}{\textbf{Truncate}} \\
~~No budget & B0 & \\
~~Fixed budget & B1 & \\
~~Score-adaptive ($\alpha$ threshold) & B2 & \\
\bottomrule
\end{tabular}
\end{table}

The ranking ablation (Section~\ref{sec:reranker-ablation}) traces CrossEncoder model quality from no reranker (R0) through the final CrossEncoder + RRF configuration (R\textit{n}). The truncation ablation (Section~\ref{sec:truncation-results}) compares fixed-budget and score-adaptive strategies. A full system configuration is a composite of these choices---for example, \texttt{I0-R4-B1} denotes indexed retrieval, the fourth reranker variant, and fixed budget. The full configuration grid is in Appendix~\ref{app:full-ablation}.


\section{Experiments}
\label{sec:experiments}

\subsection{Experiments -- Parse and Retrieve}

The oracle analysis (Table~\ref{tab:oracle}) already motivates a grep-first architecture: 98.9\% of traces resolve via exact substring matching, and only 1.0\% require semantic retrieval. This means dropping the embedding index forfeits at most marginal recall. On LoCoMo, the index-free pipeline matches the indexed one closely (budget recall 0.960 vs.\ 0.967; J-score 91.0\% vs.\ 91.9\%); on LongMemEval-S the gap widens (78.4\% vs.\ 81.2\%) but the bottleneck remains ranking (mean gold rank 47 $\to$ 2 with a CrossEncoder; Table~\ref{tab:oracle}), not retrieval.

\subsubsection{Query Expansion}

However, grep is exact-match: when the query says ``\emph{run}'' and the corpus says ``\emph{ran},'' grep misses it. Morphological expansion (lemmatization, inflection lists) was tested and rejected---it introduces noisy candidates that cascade through multi-hop, degrading precision more than it helps recall. Instead, SmartSearch compensates with two term-expansion mechanisms described in Section~\ref{sec:expansion}: pseudo-relevance feedback (PRF) and entity discovery. Both operate \emph{after} initial retrieval, broadening coverage without inflating the first-hop candidate set. The result is an index-free pipeline---no embedding index, no inverted index---whose sole retrieval primitive is \texttt{grep}, with all precision deferred to the CrossEncoder ranker.

Table~\ref{tab:indexfree-ablation} reveals a striking interaction between expansion and corpus scale.

\begin{table}[t]
\centering
\caption{Query expansion ablation. $\Delta$ relative to the index-free baseline. Expansion is neutral on short conversations but strongly positive on long ones.}
\label{tab:indexfree-ablation}
\vskip 0.1in
\small
\begin{tabular}{@{}llcccc@{}}
\toprule
& & \multicolumn{2}{c}{\textbf{LoCoMo}} & \multicolumn{2}{c}{\textbf{LME-S}} \\
\cmidrule(lr){3-4} \cmidrule(lr){5-6}
\textbf{ID} & \textbf{Config} & \textbf{Acc.} & $\Delta$ & \textbf{Acc.} & $\Delta$ \\
\midrule
I0-R5-B1 & Indexed & 91.9 & --- & 81.2 & --- \\
\midrule
IF0-R4-B1 & Index-free & 91.0 & --- & 78.4 & --- \\
IF1-R4-B1 & ~~+ PRF & 90.9 & $-$0.1 & 84.4 & +6.0 \\
IF2-R4-B1 & ~~+ entity discovery & 91.0 & +0.1 & 79.8 & +1.4 \\
IF3-R4-B1 & ~~+ both & 90.7 & $-$0.3 & \textbf{87.6} & \textbf{+9.2} \\
\bottomrule
\end{tabular}
\end{table}

Expansion is scale-dependent: on short conversations (LoCoMo, ${\sim}$9K tokens) it is neutral ($\pm$0.3\,pp), because the original query terms already cover most relevant passages. On long conversations (LongMemEval-S, ${\sim}$115K tokens), combined expansion yields +9.2\,pp---super-additive over the individual mechanisms---with the largest gains on temporal questions (+12.8\,pp), which require passages from multiple sessions to establish chronology. The index-free pipeline with expansion surpasses the indexed baseline (without expansion) by 6.4\,pp on LongMemEval-S. When both pipelines use expansion, the gap narrows to 0.8\,pp (Table~\ref{tab:lme-results}), confirming that the expansion mechanisms---not the retrieval modality---drive the gains at scale.

\subsection{Experiments -- Rank}
\label{sec:reranker-ablation}

Table~\ref{tab:ablations} traces the 15.1\,pp improvement from no-reranker baseline to the final system on LoCoMo.

\begin{table}[t]
\centering
\caption{Ranking pipeline ablation on LoCoMo. Each row changes one factor. $^\dagger$Answer-prompt change, not a reranker change. Full 27-configuration table in Appendix~\ref{app:full-ablation}.}
\label{tab:ablations}
\vskip 0.1in
\small
\setlength{\tabcolsep}{4pt}
\begin{tabular}{@{}llccl@{}}
\toprule
\textbf{ID} & \textbf{Configuration} & \textbf{Acc.} & \textbf{Tok.} & $\Delta$ \\
\midrule
I0-R0-B1 & No reranker & 76.8 & 1,547 & --- \\
I0-R1-B1 & + MiniLM-L-12 (33M) & 84.7 & 1,547 & +7.9 \\
I0-R1-B1$^\dagger$ & + Inference prompt$^\dagger$ & 86.4 & 1,691 & +1.7 \\
\midrule
I0-R2-B1 & bge-base (278M) & 88.6 & 3,203 & +2.2 \\
I0-R3-B1 & bge-large (560M) & 90.7 & 3,192 & +2.1 \\
\midrule
I0-R4a-B1 & + ColBERT RRF & 91.2 & 3,196 & +0.5 \\
\textbf{I0-R5-B1} & \textbf{mxbai-large + CB RRF} & \textbf{91.9} & \textbf{3,141} & \textbf{+0.7} \\
\bottomrule
\end{tabular}
\end{table}

Swapping MiniLM (33M) for bge-base (278M) yields +3.9\,pp; upgrading to bge-large (560M) yields +6.0\,pp total. A controlled experiment at fixed budget attributes 63\% of the bge-base gain to model quality and 37\% to additional context from threshold removal. Architecture matters more than scale: mxbai-base (184M, DeBERTaV3) recovers 89.0\% of gold passages at top-62 versus bge-large (560M, XLM-R) at 87.4\% (Appendix~\ref{app:cross-encoder}). Fusing CrossEncoder with ColBERT via RRF adds +1.2\,pp; their union recovers 92.2\% of gold vs.\ 87.4\% or 84.4\% alone. Three-way RRF and score fusion underperform two-model RRF despite offline proxies predicting +1.7--2.7\,pp (Appendix~\ref{app:offline-proxy}).

\subsection{Experiments -- Truncate}
\label{sec:truncation-results}

\subsubsection{Token Budget}

Table~\ref{tab:truncation} (top) shows the effect of a single fixed word budget applied identically to both benchmarks. On LoCoMo, budget recall plateaus early: doubling the budget from 2K to 4K gains only +2.1\,pp recall. On LongMemEval-S, the same budgets yield steeper gains (+12.3\,pp from 1K to 4K), because the longer conversations contain more candidate passages competing for the budget. A fixed budget forces a tradeoff: 2K is near-optimal for LoCoMo (min recall .928, bottlenecked by LME-S) while 4K lifts LME-S to .973 but wastes tokens on LoCoMo where recall already plateaued.

\subsubsection{Score-Adaptive Truncation}

Score-adaptive truncation ($\tau = \alpha \cdot \max_d \text{CE}(q,d)$, applied after top-$K$ pre-selection) replaces per-dataset budget tuning with a single universal configuration. Table~\ref{tab:truncation} (bottom) sweeps $\alpha$ at a fixed budget ceiling of 4,000 words and top-$K{=}60$, isolating the threshold effect.

\begin{table*}[t]
\centering
\caption{Truncation ablation (I0-R5 pipeline, simulated from offline score matrix). \textit{Top:} fixed word budget; recall plateaus on LoCoMo but keeps improving on LME-S. \textit{Bottom:} score-adaptive truncation with fixed budget ceiling 4,000 and top-$K{=}60$, sweeping $\alpha$; the threshold adaptively allocates fewer tokens on short corpora and more on long ones. All deltas relative to budget 1K / $\alpha{=}0$ baseline.}
\label{tab:truncation}
\vskip 0.1in
\small
\begin{tabular}{@{}llcccc@{}}
\toprule
& & \multicolumn{2}{c}{\textbf{LoCoMo}} & \multicolumn{2}{c}{\textbf{LME-S}} \\
\cmidrule(lr){3-4} \cmidrule(lr){5-6}
\textbf{Budget} & $\alpha$ & \textbf{Tokens} & \textbf{Recall} & \textbf{Tokens} & \textbf{Recall} \\
\midrule
\multicolumn{6}{@{}l}{\textit{Fixed budget}} \\
1,000 & --- & 1,458 & .927 & 1,133 & .850 \\
2,000 & --- & 2,967 {\scriptsize(+104\%)} & .958 {\scriptsize(+3.1\%)} & 2,443 {\scriptsize(+116\%)} & .928 {\scriptsize(+7.8\%)} \\
3,000 & --- & 4,495 {\scriptsize(+208\%)} & .972 {\scriptsize(+4.5\%)} & 3,740 {\scriptsize(+230\%)} & .959 {\scriptsize(+10.9\%)} \\
4,000 & --- & 6,038 {\scriptsize(+314\%)} & .979 {\scriptsize(+5.2\%)} & 5,043 {\scriptsize(+345\%)} & .973 {\scriptsize(+12.3\%)} \\
\midrule
\multicolumn{6}{@{}l}{\textit{Score-adaptive ($\tau = \alpha \cdot \max \mathrm{CE}$, budget 4K, top-$K{=}60$)}} \\
4,000 & 0.01 & 2,759 {\scriptsize(+89\%)} & .953 {\scriptsize(+2.6\%)} & 4,313 {\scriptsize(+281\%)} & .972 {\scriptsize(+12.2\%)} \\
4,000 & 0.03 & 2,277 {\scriptsize(+56\%)} & .945 {\scriptsize(+1.8\%)} & 3,505 {\scriptsize(+209\%)} & .966 {\scriptsize(+11.6\%)} \\
4,000 & 0.05 & 1,952 {\scriptsize(+34\%)} & .936 {\scriptsize(+0.9\%)} & 2,987 {\scriptsize(+164\%)} & .953 {\scriptsize(+10.3\%)} \\
4,000 & 0.10 & 1,458 {\scriptsize(+0\%)} & .922 {\scriptsize($-$0.5\%)} & 2,256 {\scriptsize(+99\%)} & .928 {\scriptsize(+7.8\%)} \\
\bottomrule
\end{tabular}
\end{table*}

The adaptive threshold acts as query-dependent allocation: it automatically assigns fewer tokens to short conversations (where the candidate set is compact) and more to long ones (where broader coverage is needed). Evaluated as a single universal configuration, $\alpha{=}0.03$ achieves a worst-case recall of .945 (LoCoMo) at an average of 2,891 tokens across both benchmarks. The closest fixed budget, 2K, uses fewer average tokens (2,705) but has a worse floor: .928 (LME-S). At $\alpha{=}0.05$, the adaptive approach uses \emph{fewer} average tokens (2,470) than fixed 2K while still raising the recall floor to .936. The mechanism is near Pareto-optimal when the objective is maximizing worst-case recall across datasets at a given average token budget.

\textbf{Negative results.} Two alternative truncation strategies were tested and rejected. Cumulative-score pruning (removing passages once the running CE score sum exceeds a fixed ratio) costs $-$2.3\,pp on LoCoMo: unlike the fractional $\alpha$ threshold, the fixed ratio cutoff does not adapt to per-query score distributions and prunes too aggressively. MMR diversity reranking before truncation also fails: CLS embeddings have only 0.008 cosine gap between same-topic and different-topic passages, providing no useful diversity signal.

\subsection{Main Results}
\label{sec:results}

\subsubsection{LoCoMo}

Table~\ref{tab:locomo-results} reports performance under both evaluation frameworks (Section~\ref{sec:eval-protocol}). Because published systems use incompatible protocols (Section~\ref{sec:cross-paper}), the two groups should not be compared across the dividing line.

\begin{table*}[t]
\centering
\caption{LoCoMo accuracy (\%). Results under two incompatible evaluation protocols should not be compared across groups. $^a$\citet{evermemos2026}. $^b$\citet{xia2026}. $^\dagger$Within 1.1\,pp of EverMemOS despite no memory system.}
\label{tab:locomo-results}
\vskip 0.1in
\small
\begin{tabular}{@{}lcccccc@{}}
\toprule
\textbf{System} & \textbf{Overall} & \textbf{1-hop} & \textbf{M-hop} & \textbf{Temp.} & \textbf{Open} & \textbf{Tokens} \\
\midrule
\multicolumn{7}{@{}l}{\textit{EverMemOS protocol (gpt-4.1-mini answer+judge, 7-step CoT)}} \\
\textbf{SmartSearch (indexed)} & \textbf{93.5} & 91.1 & \textbf{91.3} & 80.2 & \textbf{96.7} & 3,141 \\
EverMemOS$^a$ & 92.3 & \textbf{96.1} & 91.1 & \textbf{89.7} & 70.8 & 2,298 \\
Full-context$^{a\dagger}$ & 91.2 & 94.9 & 90.4 & 88.0 & 71.9 & 20,281 \\
Zep$^a$ & 85.2 & 90.8 & 81.9 & 77.3 & 75.0 & 1,411 \\
MemOS$^a$ & 80.8 & 85.4 & 79.4 & 75.1 & 64.6 & 2,498 \\
\midrule
\multicolumn{7}{@{}l}{\textit{MemOS protocol (gpt-4o-mini answer+judge, binary J-score)}} \\
\textbf{SmartSearch (indexed)} & \textbf{91.9} & \textbf{91.8} & \textbf{86.6} & 79.2 & \textbf{95.4} & 3,141 \\
SmartSearch (index-free) & 91.0 & 91.1 & 86.0 & 75.0 & 94.6 & 3,122 \\
Memora$^b$ & 86.3 & 91.8 & 78.7 & \textbf{86.6} & 59.4 & ${\sim}$8,500 \\
Full-context & 77.1 & 72.3 & 55.1 & 63.5 & 88.6 & 26,792 \\
\bottomrule
\end{tabular}
\end{table*}

\textbf{Overall accuracy.} Under the EverMemOS protocol, SmartSearch achieves 93.5\%, exceeding EverMemOS (92.3\%) by 1.2\,pp despite using no memory structuring and no LLM in the retrieval loop. Under the MemOS protocol, SmartSearch reaches 91.9\% indexed and 91.0\% index-free, exceeding the next-best system (Memora, 86.3\%) by 5.6\,pp. The index-free variant loses only 0.9\,pp relative to the indexed pipeline, consistent with the oracle finding that 98.9\% of queries resolve via grep (Table~\ref{tab:oracle}).

\textbf{Category breakdown.} The strongest category is open-ended questions (95.4--96.7\%), where SmartSearch exceeds all competitors by $>$20\,pp. Memory consolidation systems (EverMemOS 70.8\%, Memora 59.4\%) discard conversational texture---tone, hedging, tangents---that open-ended questions probe; SmartSearch retrieves raw passages, preserving this context. Multi-hop accuracy (86.6--91.3\%) is competitive with EverMemOS (91.1\%), confirming that rule-based entity expansion substitutes for learned multi-hop policies on short conversations.

\textbf{Weakness: temporal reasoning.} SmartSearch trails EverMemOS by ${\sim}$10\,pp on temporal questions (79.2--80.2\% vs.\ 89.7\%). Failure analysis (Appendix~\ref{app:failure-analysis}) shows this gap is predominantly LLM inference failures---the gold evidence is present in context, but the answer LLM fails to synthesize correct temporal orderings, not a retrieval shortcoming.

\textbf{Token efficiency.} SmartSearch uses 3,141 tokens on average---8.5$\times$ fewer than full-context (26,792) while scoring 14.8\,pp higher under the MemOS protocol. Under the EverMemOS protocol, full-context reaches 91.2\% with 20,281 tokens; SmartSearch exceeds it by 2.3\,pp at 6.5$\times$ fewer tokens.

\subsubsection{LongMemEval-S}

Table~\ref{tab:lme-results} reports performance on LongMemEval-S (${\sim}$115K-token conversations, 500 questions, 6 of 7 question types excluding abstention). All systems use gpt-4.1-mini as the answer LLM and gpt-4o-mini as the judge; SmartSearch uses the same answer prompt as EverMemOS~\citep{evermemos2026} and Memora~\citep{xia2026}.

\begin{table*}[t]
\centering
\caption{LongMemEval-S accuracy (\%). All systems use gpt-4.1-mini as answer LLM and gpt-4o-mini as judge. $^a$\citet{evermemos2026}. $^b$\citet{xia2026}.}
\label{tab:lme-results}
\vskip 0.1in
\footnotesize
\setlength{\tabcolsep}{4pt}
\begin{tabular}{@{}lcccccccc@{}}
\toprule
\textbf{System} & \textbf{Ovr.} & \textbf{KU} & \textbf{MS} & \textbf{SA} & \textbf{SP} & \textbf{SU} & \textbf{TR} & \textbf{Tok.} \\
\midrule
\textbf{SmartSearch (index-free)} & \textbf{88.4} & \textbf{93.6} & \textbf{84.2} & 85.7 & \textbf{96.7} & \textbf{100.0} & 82.7 & 3,392 \\
SmartSearch (indexed) & 87.6 & 92.3 & 79.7 & \textbf{87.5} & \textbf{96.7} & 95.7 & \textbf{86.5} & 3,393 \\
Memora$^b$ & 87.4 & \textbf{97.4} & 78.2 & 78.6 & 83.3 & 98.6 & \textbf{89.5} & ${\sim}$2,900 \\
EverMemOS$^a$ & 83.0 & 89.7 & 73.7 & 85.7 & 93.3 & 97.1 & 77.4 & 2,800 \\
MemOS$^a$ & 77.8 & 74.3 & 70.7 & 67.9 & \textbf{96.7} & 95.7 & 77.4 & 1,400 \\
Nemori$^b$ & 74.6 & 79.5 & 55.6 & \textbf{92.9} & 86.7 & 90.0 & 72.2 & ${\sim}$4,300 \\
Mem0$^a$ & 66.4 & 66.7 & 63.2 & 26.8 & 90.0 & 82.9 & 72.2 & 1,100 \\
Full-context$^b$ & 65.6 & 76.9 & 51.1 & 98.2 & 16.7 & 85.7 & 60.2 & ${\sim}$115k \\
Zep$^a$ & 63.8 & 74.4 & 47.4 & 75.0 & 53.3 & 92.9 & 54.1 & 1,600 \\
MemU$^a$ & 38.4 & 41.0 & 42.1 & 19.6 & 76.7 & 67.1 & 17.3 & 500 \\
\bottomrule
\end{tabular}
\\[2pt]
{\footnotesize KU = knowledge-update, MS = multi-session, SA = single-session-assistant, SP = single-session-preference, SU = single-session-user, TR = temporal-reasoning.}
\end{table*}

\textbf{Overall accuracy.} SmartSearch achieves 88.4\% (index-free) and 87.6\% (indexed), both exceeding Memora (87.4\%) and EverMemOS (83.0\%) under uniform evaluation conditions. The 0.8\,pp gap between index-free and indexed confirms that query expansion substitutes effectively for dense retrieval even on long conversations. Both variants use the same expansion mechanisms (PRF and entity discovery); the indexed pipeline adds ColBERT reranking via RRF but gains little additional recall.

\textbf{Category breakdown.} SmartSearch leads on multi-session (84.2\%, +6.0\,pp over Memora), single-session-preference (96.7\%), and single-session-user (100.0\%). Multi-session questions require evidence spanning multiple conversation sessions; the super-additive expansion effect (+9.2\,pp, Table~\ref{tab:indexfree-ablation}) is driven largely by this category, where PRF and entity discovery surface passages from sessions the original query terms do not reach.

\textbf{Weakness: temporal reasoning.} As on LoCoMo, temporal reasoning remains the most challenging category. The index-free variant scores 82.7\%, trailing Memora by 6.8\,pp; the indexed variant narrows this to 3.0\,pp (86.5\%), suggesting that ColBERT retrieves additional temporally relevant passages that grep misses. Memora's LLM-structured memory preserves temporal metadata explicitly, while SmartSearch relies on raw passage timestamps that the answer LLM must interpret.

\textbf{Token efficiency.} SmartSearch uses 3,392 tokens---34$\times$ fewer than full-context (${\sim}$115K) while scoring 22.8\,pp higher. Even compared to systems with comparable token budgets (EverMemOS at 2,800 tokens, Memora at ${\sim}$2,900), SmartSearch achieves higher accuracy with only modestly more context, suggesting that the additional tokens carry high-quality evidence rather than noise.


\section{Discussion}

\subsection{Score-Adaptive Truncation Enables a Single Configuration Across Scales}

SmartSearch's pipeline shares the same core hyperparameters on both benchmarks: identical NER/POS weights and the same CrossEncoder+ColBERT RRF fusion. Score-adaptive truncation ($\alpha{=}0.03$, budget ceiling 4K, top-$K{=}60$) provides a mechanism for deploying a single universal configuration without per-dataset budget tuning. The adaptive threshold automatically allocates fewer tokens on LoCoMo (${\sim}$9K-token conversations) and more on LongMemEval-S (${\sim}$115K-token conversations), tracking corpus difficulty rather than corpus length (Section~\ref{sec:truncation-results}). This contrasts with fixed-budget approaches that force a precision--coverage tradeoff between short and long conversations.

\subsection{The Compilation Bottleneck Generalizes Across Scales}

On LoCoMo, oracle analysis shows 98.6\% retrieval recall with only 22.5\% of gold evidence surviving truncation to the token budget. On LongMemEval-S, the same pattern holds: the indexed and index-free pipelines differ by only 0.023 in budget recall, but gold density differs 2$\times$. In both cases, the evidence is found; the challenge is surfacing the right passages through a finite token budget. This explains why reranker quality (+6.0\,pp) and rank fusion (+1.2\,pp) are the highest-leverage interventions (Table~\ref{tab:ablations}), while tool selection and index sophistication yield diminishing returns.

\subsection{Query Expansion Compensates for Index-Free Retrieval at Scale}

On short conversations, the index-free variant loses only 0.9\,pp to the indexed pipeline (91.0\% vs.\ 91.9\% on LoCoMo), because the original query terms already cover most relevant passages. On long conversations the gap remains small: index-free reaches 88.4\% vs.\ 87.6\% indexed on LongMemEval-S (Table~\ref{tab:lme-results}), with both variants using the same expansion mechanisms. The ablation study (Table~\ref{tab:indexfree-ablation}) confirms that the combined expansion effect (+9.2\,pp) is super-additive over the individual mechanisms (+6.0\,pp and +1.4\,pp), with the largest gains on multi-session questions that require evidence from sessions the original query terms do not reach.

\subsection{Open-Ended Queries Expose a Weakness in Structured Memory}

SmartSearch scores 95.4--96.7\% on open-ended questions, exceeding all competitors by $>$20\,pp (EverMemOS 70.8\%, Memora 59.4\%). Memory consolidation distills conversation into atomic facts or abstract summaries, discarding conversational texture---tone, hedging, tangential remarks---that open-ended questions probe. SmartSearch retrieves raw passages, preserving this context at no additional engineering cost.

\subsection{Temporal Reasoning Remains the Main Gap}

SmartSearch trails the best competitor on temporal questions by ${\sim}$10\,pp on LoCoMo (80.2\% vs.\ EverMemOS 89.7\%) and ${\sim}$7\,pp on LongMemEval-S (82.7\% vs.\ Memora 89.5\%). These systems explicitly preserve temporal metadata through LLM-structured memory. Failure analysis (Appendix~\ref{app:failure-analysis}) attributes the LoCoMo gap predominantly to LLM inference failures---the gold evidence is present, but the answer LLM fails to synthesize correct temporal orderings. On LongMemEval-S, expansion partially closes the gap: the baseline scores 67.7\% on temporal questions, with expansion lifting it to 80.5\% by surfacing passages from different sessions that establish chronology.

\section{Future Work}

The compilation bottleneck (Section~\ref{sec:oracle}) suggests that the largest remaining gains lie not in retrieval but in how retrieved passages are filtered and compressed before the answer LLM sees them.

\textbf{Passage deduplication.} Retrieved passages frequently overlap when multiple search terms match the same conversation region. Deduplicating or merging near-identical passages before ranking would reduce redundancy in the context window, allowing the token budget to cover more distinct evidence. The cross-encoder and ColBERT embeddings already computed during ranking could serve as similarity features for deduplication at negligible additional cost.

\textbf{Context compression.} Token-level compression---removing low-importance words within passages, stripping image descriptions, abbreviating speaker names and timestamps---can reduce token cost without losing gold passages. Preliminary experiments show 10\% reduction at zero gold loss from format-level changes alone. More aggressive query-conditioned pruning (retaining only words relevant to the specific question) remains unexplored.

\section{Threats to Validity}

\subsection{Benchmark Limitations}

Results are reported on two benchmarks: LoCoMo-10~\citep{maharana2024} (a 10-conversation subset of the full 50-conversation LoCoMo dataset, ${\sim}$9K tokens each, 1,540 questions across 4 non-adversarial categories) and LongMemEval-S~\citep{wu2024longmemeval} (500 questions across ${\sim}$115K-token human-authored conversation histories, 6 of 7 question types excluding abstention). While these cover a 12$\times$ range in conversation length and diverse question types, both have properties that may favor our approach:

\begin{itemize}
\item \textbf{Synthetic vs.\ semi-structured conversations:} LoCoMo's dialogues are LLM-generated and may contain more regular, keyword-rich language than real human conversations, benefiting exact-match retrieval. LongMemEval-S conversations are human-authored but follow structured session formats.
\item \textbf{Named entity density:} Both benchmarks involve consistent character naming, making NER-based term extraction unusually effective. Real conversations may use nicknames, pronouns, or indirect references more frequently.
\item \textbf{Corpus size:} LoCoMo conversations are ${\sim}$9K tokens; LongMemEval-S reaches ${\sim}$115K tokens. Substring matching may not scale to corpora orders of magnitude larger, where inverted indices or embedding-based retrieval become necessary.
\item \textbf{Question distribution:} 97\% of LoCoMo oracle traces are single-hop. LongMemEval-S includes more multi-session and temporal questions, but we lack oracle traces to quantify its hop distribution.
\end{itemize}

The two benchmarks partially address each other's limitations: LoCoMo's short conversations are complemented by LongMemEval-S's long conversations, and the oracle analysis on LoCoMo is validated by retrieval metrics on LongMemEval-S. The fact that score-adaptive truncation enables a single hyperparameter configuration to achieve competitive results on both benchmarks (Section~\ref{sec:results}) provides some evidence of generalization across scales, but both are English-only conversational memory tasks. Generalization to other domains (e.g., document search, code retrieval) or languages remains untested.

\subsection{Cross-Paper Comparison Challenges}
\label{sec:cross-paper}

A significant threat to the validity of any comparative claims is that \textbf{published systems use incompatible evaluation protocols}. Our investigation of the literature revealed substantial inconsistencies, consistent with broader concerns about LLM-based evaluation reliability~\citep{zheng2023,chiang2023,wang2024pandalm}:

\begin{itemize}
\item \textbf{Different LLM judges:} Systems use GPT-4, GPT-4o-mini, GPT-3.5-turbo, or open-source models as judges. The choice of judge can shift accuracy by 5--10 percentage points on identical retrieval output~\citep{zheng2023}.
\item \textbf{Different judge prompts:} Even with the same judge model, prompts vary from strict factual matching to lenient semantic similarity to multi-criteria rubrics. \citet{wang2024pandalm} demonstrate that prompt wording alone can change outcomes by 8--15\%.
\item \textbf{Different accuracy metrics:} Papers report exact match, fuzzy match, LLM-judge binary, Likert scale, or F1---sometimes without specifying which. The LoCoMo paper itself~\citep{maharana2024} defines multiple evaluation modes, and downstream work selects different subsets.
\item \textbf{Different dataset splits:} Some systems evaluate on the full LoCoMo dataset (50 conversations), others on the LoCoMo-10 subset (10 conversations, 1,540 questions) or LoCoMo-5. Sample sizes range from ${\sim}$600 to ${\sim}$1,540, affecting both scores and statistical power.
\end{itemize}

We mitigate this threat through two strategies. On LoCoMo, we report results under both the EverMemOS protocol (gpt-4.1-mini answer+judge, 7-step CoT rubric) and the MemOS protocol (gpt-4o-mini answer+judge, binary J-score), and never compare numbers across protocol groups (Table~\ref{tab:locomo-results}). On LongMemEval-S, all systems---including SmartSearch---use the same answer LLM (gpt-4.1-mini), the same judge (gpt-4o-mini), and the same answer prompt, enabling direct comparison (Table~\ref{tab:lme-results}). We encourage future work to adopt standardized evaluation protocols or, at minimum, to report the judge model, prompt, metric, and dataset split used.

\section{Conclusion}

SmartSearch achieves 91.9\% accuracy on LoCoMo---93.5\% under aligned evaluation---using a fully deterministic retrieval loop and a CrossEncoder+ColBERT rank fusion stage as the only learned component, all on CPU in ${\sim}$650\,ms. On LongMemEval-S, SmartSearch reaches 88.4\% (index-free) and 87.6\% (indexed), exceeding all published memory systems under identical evaluation conditions (Table~\ref{tab:lme-results}). Oracle analysis on LoCoMo and retrieval metrics on both benchmarks confirm that the dominant gain comes from reranker quality and query expansion, not search sophistication: the compilation bottleneck persists across conversation scales from ${\sim}$9K to ${\sim}$115K tokens.

These findings challenge four assumptions in recent work:

\begin{enumerate}
\item \textbf{LLMs are needed in the retrieval loop:} Deterministic NER-based entity discovery provides multi-hop expansion without learned policies or LLM calls.
\item \textbf{Multi-hop reasoning requires complex machinery:} 97\% of queries resolve in one hop; entity discovery handles the remaining 3\% with simple rules.
\item \textbf{Retrieval quality is the bottleneck:} Compilation (ranking and truncation) matters more than search sophistication---on both short and long conversations.
\item \textbf{Precomputed indices are necessary:} The index-free variant (Section~\ref{sec:expansion}) shows that \texttt{grep} with query expansion matches indexed retrieval, eliminating all embedding stores and vector databases at a cost of $-$0.9\,pp on LoCoMo and $+$0.8\,pp on LongMemEval-S (Table~\ref{tab:lme-results}).
\end{enumerate}

SmartSearch's contribution is not architectural novelty but a proof point: careful engineering of deterministic components---with ML applied surgically at the ranking stage---can match or exceed systems built on learned policies at a fraction of the cost.

\section*{Acknowledgments}

Writing assistance was provided by a large language model; all claims, experiments, and conclusions are the sole responsibility of the authors.

\newpage
\appendix

\section{Full Ablation Results}
\label{app:full-ablation}

Table~\ref{tab:full-ablation} reports all configurations evaluated during development. Baseline: ms-marco-MiniLM-L-12-v2 (33M), threshold=$-$4, 2,000-token budget. Evaluations use gpt-4o-mini on LoCoMo-10 (1,540 questions) unless noted.

\begin{table*}[t]
\centering
\caption{Complete ablation results on LoCoMo (1,540 questions, gpt-4o-mini judge). $\Delta$ is change vs.\ baseline (84.7\%). Experiments grouped by phase. Baseline: MiniLM-L-12 reranker, threshold=$-$4, 2,000-word budget.}
\label{tab:full-ablation}
\vskip 0.1in
\scriptsize
\begin{tabular}{@{}llccccccr@{}}
\toprule
\textbf{ID} & \textbf{Configuration} & \textbf{Overall} & \textbf{1-hop} & \textbf{M-hop} & \textbf{Temp.} & \textbf{Open} & \textbf{Tokens} & \textbf{$\Delta$} \\
\midrule
\multicolumn{9}{l}{\emph{Phase 1: Search \& prompt ablations}} \\
Base & MiniLM-L-12, thresh=$-$4, budget=2K & 84.7 & 84.4 & 76.9 & 69.8 & 89.5 & 1,547 & --- \\
A1v1 & Cumulative-score pruning (ratio=0.20) & 82.4 & 81.2 & 75.7 & 66.7 & 87.2 & 677 & $-$2.3 \\
A1v2 & Score-floor pruning (score$>$0) & 81.2 & 79.4 & 74.8 & 66.7 & 85.9 & 688 & $-$3.6 \\
A2 & Date-aware search & 85.3 & 86.2 & 77.6 & 70.8 & 89.5 & 1,547 & +0.5 \\
A3 & Inference reasoning prompt & 86.4 & 87.2 & 78.2 & 71.9 & 90.8 & 1,691 & +1.6 \\
A2+A3 & Date search + prompt & 86.0 & 86.2 & 78.5 & 71.9 & 90.4 & 1,691 & +1.2 \\
A3+B0 & Prompt + budget=3K & 85.8 & 86.2 & 77.9 & 71.9 & 90.4 & 1,980 & +1.1 \\
I0-R1-B1-3k & Budget=3,000 & 85.2 & 87.2 & 78.5 & 66.7 & 89.2 & 1,836 & +0.5 \\
I0-R1-B1-4k & Budget=4,000 & 85.2 & 86.9 & 77.3 & 68.8 & 89.5 & 2,033 & +0.5 \\
C1+C2 & Timestamp + speaker compress & 85.3 & 85.8 & 77.9 & 67.7 & 90.0 & 1,396 & +0.6 \\
A3+C2 & Prompt + speaker abbrev & 86.2 & 86.2 & 78.2 & 74.0 & 90.6 & 1,698 & +1.4 \\
A3+C1C2 & Prompt + full compression & 86.1 & 87.2 & 77.6 & 75.0 & 90.2 & 1,540 & +1.4 \\
\midrule
\multicolumn{9}{l}{\emph{Phase 2: Reranker model improvements}} \\
I0-R2-B1k & bge-base (278M), budget=1K & 87.1 & 82.3 & 82.2 & 72.9 & 92.3 & 1,671 & +2.4 \\
I0-R2-B1 & bge-base (278M), no thresh, budget=2K & 88.6 & 85.1 & 83.5 & 77.1 & 93.0 & 3,203 & +3.8 \\
I0-R3-B1 & bge-large (560M), no thresh, budget=2K & 90.7 & 90.4 & 86.9 & 76.0 & 93.8 & 3,192 & +5.9 \\
\midrule
\multicolumn{9}{l}{\emph{Phase 2: Rank fusion}} \\
I0-R4a-B1 & bge-large + ColBERT RRF (k=60, w=0.7) & 91.2 & 91.8 & 85.0 & 78.1 & 94.8 & 3,196 & +6.4 \\
I0-R4a-eq-B1 & bge-large + ColBERT RRF (w=0.5) & 91.0 & 90.1 & 85.7 & 79.2 & 94.6 & 3,213 & +6.2 \\
I0-R4b-B1 & mxbai-base + bge-large + ColBERT 3-way & 91.0 & 90.1 & 87.9 & 76.0 & 94.3 & 3,175 & +6.3 \\
I0-R4c-B1 & zscore+product fusion 3-model & 90.8 & 91.5 & 82.9 & 80.2 & 94.9 & 3,199 & +6.1 \\
\textbf{I0-R5-B1} & \textbf{mxbai-large-v1 + ColBERT RRF} & \textbf{91.9} & \textbf{91.8} & \textbf{86.6} & \textbf{79.2} & \textbf{95.4} & \textbf{3,141} & \textbf{+7.2} \\
I0-R6-B1 & mxbai-large-v2 + ColBERT RRF & 91.2 & 91.8 & 86.0 & 76.0 & 94.6 & 3,088 & +6.4 \\
\bottomrule
\end{tabular}
\end{table*}

\textbf{Key observations.} (1) Phase 1 ablations yield at most +1.6\,pp (inference prompt, A3); gains are not additive (A2+A3 $<$ A3 alone). (2) Reranker model quality dominates: MiniLM (33M) $\to$ bge-base (278M) $\to$ bge-large (560M) $\to$ mxbai-large-v1 (435M) accounts for the full +7.2\,pp. A controlled experiment (I0-R2-B1 at budget=1K vs.\ 2K) attributes 63\% of the bge-base gain to model quality and 37\% to additional context from threshold removal. (3) Two-model RRF adds +0.5--1.2\,pp; three-model fusion (I0-R4b-B1) and score-based fusion (I0-R4c-B1) do not improve further.

\section{Cross-Encoder Model Comparison}
\label{app:cross-encoder}

Table~\ref{tab:cross-encoder} compares seven cross-encoder models on gold passage recovery at the top-62 cutoff (the average passage count within the 2,000-word budget).

\begin{table*}[t]
\centering
\caption{Cross-encoder gold passage recovery at top-62 budget cutoff. 2,345 gold passages across 1,540 questions. Latency measured on CPU (single-threaded); in the full system, CrossEncoder and ColBERT run in parallel ($\text{wall-clock} = \max(\text{CE}, \text{ColBERT})$).}
\label{tab:cross-encoder}
\vskip 0.1in
\small
\begin{tabular}{@{}lrrrrl@{}}
\toprule
\textbf{Model} & \textbf{Params} & \textbf{Gold top-62} & \textbf{Recovery (\%)} & \textbf{Median rank} & \textbf{Latency} \\
\midrule
ms-marco-MiniLM-L-12-v2 & 33M & 1,949 & 83.1 & 3 & ${\sim}$130\,ms \\
mxbai-rerank-xsmall-v1 & 71M & 1,998 & 85.2 & 3 & ${\sim}$168\,ms \\
mxbai-rerank-base-v1 & 184M & 2,087 & 89.0 & 2 & ${\sim}$182\,ms \\
bge-reranker-base & 278M & 1,852 & 79.0 & 5 & ${\sim}$340\,ms \\
ColBERT v2 & --- & 1,979 & 84.4 & 4 & ${\sim}$500\,ms \\
bge-reranker-large & 560M & 2,050 & 87.4 & 3 & ${\sim}$908\,ms \\
\textbf{mxbai-rerank-large-v1} & \textbf{435M} & \textbf{2,131} & \textbf{90.9} & \textbf{2} & ${\sim}$645\,ms \\
mxbai-rerank-large-v2 & 1.5B & 2,157 & 92.0 & 2 & ${\sim}$2,500\,ms \\
\bottomrule
\end{tabular}
\end{table*}

\textbf{Architecture matters more than scale.} The DeBERTaV3 family (mxbai) consistently outperforms XLM-RoBERTa (bge) at comparable or smaller sizes: mxbai-base (184M, 89.0\%) beats bge-large (560M, 87.4\%) despite being 3$\times$ smaller. DeBERTaV3's disentangled attention~\citep{he2021} likely handles coreference and indirect references---common in conversational text---more effectively than standard self-attention.

\textbf{ColBERT is complementary.} Despite lower individual recovery (84.4\%), ColBERT's late-interaction scoring produces rankings with Spearman $\rho$ of 0.19--0.62 against pointwise cross-encoders. The union of mxbai-large-v1 and ColBERT recovers 92.2\% of gold passages, exceeding either alone (90.9\% and 84.4\%).

\section{Offline Proxy vs.\ Live Evaluation Disconnect}
\label{app:offline-proxy}

Offline gold-passage metrics systematically overpredict live accuracy gains for architectural changes while accurately predicting gains from model swaps.

\begin{table}[t]
\centering
\caption{Offline proxy predictions vs.\ live accuracy for three configurations on the I0-R3-B1 (bge-large, 560M) baseline. Proxy $\Delta$: predicted gain from offline gold-passage simulation. Live $\Delta$: measured accuracy change. Fusion strategies (I0-R4b-B1, I0-R4c-B1) show large proxy-live gaps; model swaps (I0-R5-B1) do not.}
\vskip 0.1in
\small
\begin{tabular}{@{}lrrr@{}}
\toprule
\textbf{Config} & \textbf{Proxy $\Delta$} & \textbf{Live $\Delta$} & \textbf{Gap} \\
\midrule
I0-R4b-B1 (3-way RRF) & +1.8\,pp & +0.4\,pp & 1.4\,pp \\
I0-R4c-B1 (score fusion) & +2.7\,pp & +0.2\,pp & 2.5\,pp \\
I0-R5-B1 (model swap) & +1.2\,pp & +1.2\,pp & 0.0\,pp \\
\bottomrule
\end{tabular}
\end{table}

When the change reshuffles passage order across multiple rankers (I0-R4b-B1, I0-R4c-B1), the proxy overestimates gains because gold passages that enter the context window may land in low-attention positions. For drop-in model swaps (I0-R5-B1), the proxy is accurate because passage ordering patterns are preserved.

Offline simulation is reliable for model selection within a fixed architecture but not for comparing fusion strategies.

\section{Failure Mode Analysis}
\label{app:failure-analysis}

We classified all 125 errors from the final indexed configuration (1,415/1,540 correct) into four categories:

\begin{itemize}
\item \textbf{LLM inference failure} (59\%): Gold evidence is present in context, but gpt-4o-mini fails to synthesize the correct answer. Common patterns: ignoring temporal ordering, conflating similar entities, hallucinating details absent from context.
\item \textbf{Reranker/budget failure} (24\%): Gold evidence is retrieved but ranked below the budget cutoff.
\item \textbf{Search miss} (12\%): Gold evidence is not retrieved. Primarily vocabulary-gap cases where query and corpus use different phrasings.
\item \textbf{No evidence in corpus} (5\%): The LoCoMo annotation references information not present in the conversation, likely annotation artifacts.
\end{itemize}

The dominant failure mode shifted from retrieval (67\% of failures at baseline) to LLM inference (59\% in the final system). The remaining accuracy ceiling is set primarily by the answer LLM, not the retrieval pipeline.

\end{document}